\documentclass[letterpaper]{article}

\makeatletter
\let\O@argtabularcr\@argtabularcr
\def\O@xtabularcr{\@ifnextchar[\O@argtabularcr{\ifnum 0=`{\fi}\cr}}
\let\O@tabacol\@tabacol
\let\O@tabclassiv\@tabclassiv
\let\O@tabclassz\@tabclassz
\let\O@tabarray\@tabarray
\def\author@tabular{\authorsize\def\@halignto{}\@authortable}
\let\endauthor@tabular=\endtabular
\def\author@tabcrone{{\ifnum0=`}\fi\O@xtabularcr\affilsize\itshape
 \let\\=\author@tabcrtwo\ignorespaces}
\def\author@tabcrtwo{{\ifnum0=`}\fi\O@xtabularcr[-3\p@]\affilsize\itshape
 \let\\=\author@tabcrtwo\ignorespaces}
\def\@authortable{\leavevmode \hbox \bgroup $\let\@acol\O@tabacol
 \let\@classz\O@tabclassz \let\@classiv\O@tabclassiv
 \let\\=\author@tabcrone \ignorespaces \O@tabarray}
\makeatother

\usepackage{aaai}
\usepackage{balance}

\usepackage[inline]{enumitem}
\usepackage{float}
\usepackage[table,xcdraw]{xcolor}
\usepackage[fleqn]{amsmath}
\usepackage{algorithm}
\usepackage{algpseudocode}
\usepackage[inline]{enumitem}
\usepackage{graphicx}
\usepackage{tikz-cd}
\usepackage{multirow} 
\usepackage{cprotect}
\usepackage{comment}
\usepackage[T1]{fontenc}
\usepackage{caption}
\usepackage{diagbox}
\usepackage{listings}
\usepackage{wrapfig}
\usepackage{stfloats}
\usepackage{slashbox}
\definecolor{darkbrown}{rgb}{0.4, 0.26, 0.13}
\definecolor{burgundy}{rgb}{0.5, 0.0, 0.13}

\begin{document}

%
\title{Knowledge-driven Natural Language Understanding\\of English Text and its Applications\thanks{Authors partially supported by NSF grants IIS 1718945, IIS 1910131, and IIP 1916206}}

\author {
  Kinjal Basu$^1$, Sarat  Varanasi$^1$, Farhad Shakerin$^1$, Joaquin Arias$^2$ and Gopal Gupta$^1$\\[.5em]
  {  \large $^1$Department of Computer Science \hfill $^2$Artificial Intelligence Research Group}\\{\large ~\hspace{.3em}The University of Texas at Dallas, USA\hspace{.7em} \hfill Universidad Rey Juan Carlos, Madrid, Spain}
}


\maketitle
\begin{abstract}
\begin{quote}
Understanding the meaning of a text is a fundamental challenge of natural language understanding (NLU) research. An ideal NLU system should process a language in a way that is not exclusive to a single task or a dataset. Keeping this in mind, we have introduced a novel knowledge driven semantic representation approach for English text. By leveraging the VerbNet lexicon, we are able to map syntax tree of the text to its commonsense meaning represented using basic knowledge primitives. The general purpose knowledge represented from our approach can be used to build any reasoning based NLU system that can also provide justification. We applied this approach to construct two NLU applications that we present here: SQuARE (Semantic-based Question Answering and Reasoning Engine) and StaCACK (Stateful Conversational Agent using Commonsense Knowledge). 
Both these systems work by ``truly understanding'' the natural language text they process and both provide natural language explanations for their responses while maintaining high accuracy. 

\end{quote}
\end{abstract}

\section{Introduction}

The long term goal of natural language understanding (NLU) research is to make applications, e.g., chatbots and question answering (QA) systems, that act exactly like a human assistant. A human assistant will understand the user's intent and fulfill the task. The task can be answering questions about a story, giving directions to a place, or  reserving a table in a restaurant by knowing user's preferences. Human level understanding of natural language is needed for an NLU application that aspires to act exactly like a human. To understand the meaning of a natural language sentence, humans first process the syntactic structure of the sentence and then infer its meaning. Also, humans use commonsense knowledge to understand the often complex and ambiguous meaning of natural language sentences. Humans interpret a passage as a sequence of sentences and will normally process the events in the story in the same order as the sentences. Once humans understand the meaning of a passage, they can answer questions posed, along with an explanation for the answer. Moreover, by using commonsense, a human assistant understands the user's intended task and asks questions to the user about the required information to successfully carry-out the task. Also, to hold a goal oriented conversation, a human remembers all the details given in the past and most of the time performs non-monotonic reasoning to accomplish the assigned task. We believe that an automated QA system or a goal oriented closed domain chatbot should work in a similar way. 

If we want to build AI systems that emulate humans, then understanding natural language sentences is the foremost priority for any NLU application. In an ideal scenario, an NLU application should map the sentence to the knowledge (semantics) it represents, augment it with commonsense knowledge related to the concepts involved--just as humans do---then use the combined knowledge to do the required reasoning. In this paper, we introduce our novel algorithm for automatically generating the semantics corresponding to each English sentence using the comprehensive verb-lexicon for English verbs - VerbNet \cite{vn}. For each English verb, VerbNet gives the syntactic and semantic patterns. The algorithm employs partial syntactic matching between parse-tree of a sentence and a verb's \textit{frame syntax} from VerbNet to obtain the meaning of the sentence in terms of VerbNet's primitive predicates.

This matching is motivated by denotational semantics of programming languages and can be thought of as mapping parse-trees of sentences to knowledge that is constructed out of semantics provided by VerbNet. The VerbNet semantics is expressed using a set of primitive predicates that can be thought of as the \textit{semantic algebra} of the denotational semantics.

We also show two applications of our approach. SQuARE, a question answering system for reading comprehension, is capable of answering various types of reasoning questions asked about a passage. SQuARE uses knowledge in the passage augmented with commonsense knowledge. Going a step further, we leverage SQuARE to build a general purpose closed-domain goal-oriented chatbot framework - StaCACK (pronounced as \textit{stack}). Our work reported here builds upon our prior work in natural language QA as well as visual QA \cite{caspr,aqua,square}.

\section{Background and Contribution}

We next describe some of the key technologies we employ. Our effort is based on answer set programming (ASP) technology \cite{asp}, specifically, its goal-directed implementation in the s(CASP) system.  ASP supports nonmonotonic reasoning through negation as failure that is crucial for modeling commonsense reasoning (via defaults, exceptions and preferences) \cite{asp}. 
We assume that the reader is familiar with ASP \cite{asp}, denotational semantics \cite{d_semantics} and English language parsers (Stanford CoreNLP \cite{corenlp} and spaCy \cite{spacy}). We give a brief overview of ASP and denotational semantics.

\medskip
\noindent\textbf{ASP:}
%
Answer Set Programming (ASP) is a declarative logic programming language that extends it with negation-as-failure. ASP is a highly expressive paradigm that can elegantly express complex reasoning methods, including those used by humans, such as default reasoning, deductive and abductive reasoning, counterfactual reasoning, constraint satisfaction~\cite{baral,asp}.
ASP supports better semantics for negation ({\it negation as failure}) than does standard logic programming and Prolog. An ASP program consists of rules that look like Prolog rules. The semantics of an ASP program {$\Pi$} is given in terms of the answer sets of the program \texttt{ground($\Pi$)}, where \texttt{ground($\Pi$)} is the program obtained from the substitution of elements of the \textit{Herbrand universe} for variables in $\Pi$~\cite{baral}.
The rules in an ASP program are of the form:
\begin{center}
  \texttt{p :- q$_1$, ..., q$_m$, not r$_1$, ..., not r$_n$.}
\end{center}
\noindent where $m \geq 0$ and $n \geq 0$. Each of \texttt{p} and \texttt{q$_i$} ($\forall i \leq m$) is a literal, and each \texttt{not r$_j$} ($\forall j \leq n$) is a \textit{naf-literal} (\texttt{not} is a logical connective called \textit{negation-as-failure} or \textit{default negation}). The literal \texttt{not r$_j$} is true if proof of {\tt r$_j$} \textit{fails}. Negation as failure allows us to take actions that are predicated on failure of a proof. Thus, the rule {\tt r :- not s.} states that {\tt r} can be inferred if we fail to prove {\tt s}. Note that in the rule above, the head literal {\tt p} is optional. A headless rule is called a constraint which states that conjunction of {\tt q$_i$}'s and \texttt{not r$_j$}'s should yield \textit{false}. 

The declarative semantics of an Answer Set Program \texttt{P} is given via the Gelfond-Lifschitz transform~\cite{baral,asp} in terms of the answer sets of the program \texttt{ground($\Pi$)}. ASP also supports classical negation. A classically negated predicate (denoted {\tt -p} means that {\tt p} is definitely false. Its definition is no different from a positive predicate, in that explicit rules have to be given to establish {\tt -p}. More details on ASP can be found elsewhere~\cite{baral,asp}. 

The goal in ASP is to compute an {\it answer set} given an answer set program, i.e., compute the set that contains all propositions that if set to true will serve as a model of the program (those propositions that are not in the set are assumed to be false).  Intuitively, the rule above says that {\tt p} is in the answer set if {\tt q$_1$, ..., q$_m$} are in the answer set and {\tt r$_1$, ..., r$_n$} are not in the answer set. 
ASP can be thought of as Prolog extended with a sound semantics of negation-as-failure that is based on the stable model semantics~\cite{stablemodel}. 

\medskip 
\noindent\textbf{s(CASP) System:}
    s(CASP) \cite{scasp}  is a query-driven, goal-directed implementation of ASP that includes constraint solving over reals. Goal-directed execution of s(CASP) is indispensable for automating commonsense reasoning, as traditional grounding and SAT-solver based implementations of ASP may not be scalable. There are three major advantages of using the s(CASP) system: (i) s(CASP) does not ground the program, which makes our framework scalable, (ii) it only explores the parts of the knowledge base that are needed to answer a query, and (iii) it provides natural language justification  (proof tree) for an answer \cite{scasp_justification}.

\medskip 
\noindent\textbf{Denotational Semantics:} 
In programming language research, denotational semantics is a widely used approach to formalize the meaning of a  programming language in terms of mathematical objects (called \textit{domains}, such as integers, truth-values, tuple of values, and, mathematical functions) \cite{d_semantics}. Denotational semantics of a programming language  has three components \cite{d_semantics}:

\begin{enumerate}
    \item \textbf{{Syntax}}: specified as abstract syntax trees.
    \item \textbf{{Semantic Algebra}}: these are the basic domains along with the associated operations; meaning of a program is expressed in terms of these basic operations applied to the elements in the domain.
    \item \textbf{{Valuation Function}}: these are mappings from abstract syntax trees (and possibly the semantic algebra) to values in the semantic algebra.
\end{enumerate} 

\noindent Given a program \textbf{P} written in language \textbf{L}, \textbf{P}'s denotation (meaning), expressed in terms of the semantic algebra, is obtained by applying the valuation function of \textbf{L} to program \textbf{P}'s syntax tree. Details can be found elsewhere \cite{d_semantics}.   

\begin{figure}[t]
\tiny
\centering
\begin{tabular}{|p{1.2cm}|p{6cm}|}
    \hline
     \multicolumn{2}{|l|}{\textbf{NP V NP}}   \\
    \hline
    \textbf{Example}  & \textbf{\textit{``She grabbed the rail''}} \\
    \hline 
    \textbf{Syntax}  & \textbf{Agent} \textbf{V} \textbf{Theme}  \\
    \hline  
    \textbf{Semantics}  & \textcolor{burgundy}{Continue(}\textbf{E,Theme}\textcolor{burgundy}{)}, \textcolor{burgundy}{Cause(}\textbf{Agent, E}\textcolor{burgundy}{)}\\ 
    & \textcolor{burgundy}{Contact(}\textbf{During(E),Agent,Theme}\textcolor{burgundy}{)} \\
    \hline
\end{tabular}
\caption{VerbNet frame instance for the verb class \textit{grab}} 
    \label{fig:verbnet-example}
\end{figure}

\medskip 

\noindent\textbf{VerbNet:} 
Inspired by Beth Levin’s classification of verbs and their syntactic alternations \cite{verbclasses}, VerbNet \cite{vn} is the largest online network of English verbs. A verb class in VerbNet is mainly expressed by \textit{syntactic frames}, \textit{thematic roles},  and  \textit{semantic representation}. The VerbNet
lexicon identifies  thematic roles and  syntactic patterns of each verb class and infers the common syntactic structure and semantic relations for all the member verbs. Figure \ref{fig:verbnet-example} shows an example of a VerbNet frame of the verb class \textit{grab}. 


This paper makes the following novel contributions: (i) it presents a domain independent English text to answer set program generator, (ii) demonstrates two robust, scalable, and interpretable applications of our semantic-driven approach, SQuARE and StaCACK, both of which
demonstrate improved performance over machine learning (ML) based systems with regards to accuracy and explainability, 
and (iii) shows how the s(CASP) query-driven ASP system is crucial for commonsense reasoning as it guarantees a correct answer if the knowledge representation is accurate.
Our work is based purely on reasoning and does not require any manual intervention other than providing (reusable) commonsense knowledge coded in ASP. It paves the way for developing advanced NLU systems based on ``truly understanding''  text or human dialog. 
 
\section{Semantics driven ASP Code Generation} 
Similar to the denotational approach for meaning representation of a programming language, an ideal NLU system should use denotational semantics to compositionally map text syntax to its meaning. Knowledge primitives should be represented using the \textit{semantic algebra} \cite{d_semantics} of  well understood concepts. Then the semantics along with the commonsense knowledge represented using the same semantic algebra can be used to construct different NLU applications, such as QA system, chatbot, information extraction system, text summarization, etc. The ambiguous nature of natural language is the main hurdle in treating it as a programming language. English is no exception and the meaning of an English word or sentence may depend on the context. The algorithm we present takes the syntactic parse tree of an English sentence and uses VerbNet to automatically map the parse tree to its denotation, i.e., the knowledge it represents.

\begin{figure}[h]
    \centering
    \includegraphics[scale = 0.8]{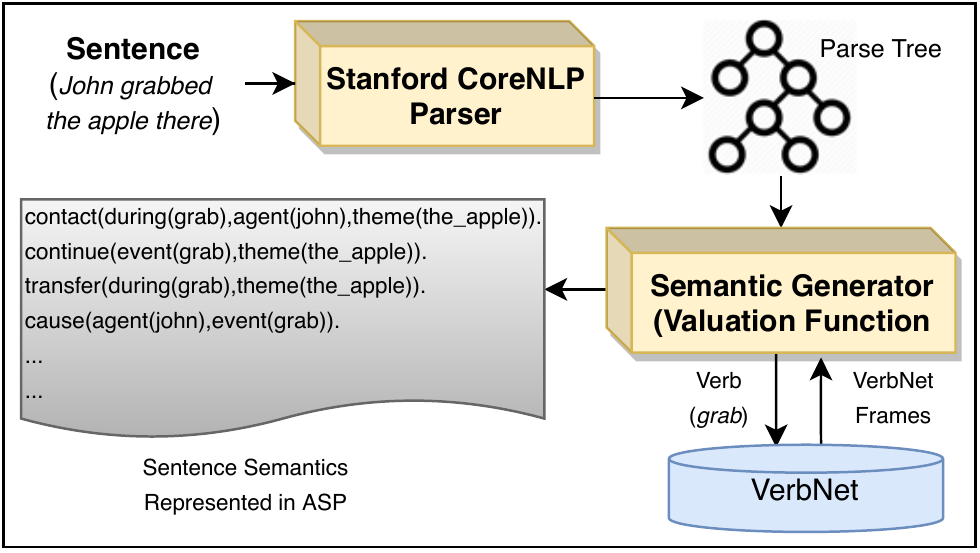}
    \caption{English to ASP translation process }
    \label{fig:semantic_algebra}
    \end{figure}

An English sentence that consists of an action verb (i.e., not a \textit{be} verb) always describes an event. The verb also constrains the relation among the event participants. VerbNet encapsulates all of this information using verb classes that represent a verb set with similar meanings. So each verb is a part of one or more classes. For each class, it provides the skeletal parse tree (frame syntax) for different usage of the verb class and the respective semantics (frame semantic). The semantic definition of each frame uses pre-defined predicates of VerbNet that have thematic-roles (AGENT, THEME, etc.) as arguments. Thus, we can imagine VerbNet as a very large valuation (semantic) function that maps syntax tree patterns to their respective meanings. As we use ASP to represent the knowledge, the algorithm generates the sentence's semantic definition in ASP. \textit{Our goal is to find the partial matching between the sentence parse tree and the VerbNet frame syntax and ground the thematic-role variables so that we can get the semantics of the sentence from the frame semantics and represent it in ASP.} 

The illustration of the process of semantic knowledge generation from a sentence is described in the Figure \ref{fig:semantic_algebra}. We have used Stanford's CoreNLP parser \cite{corenlp} to generate the parse tree, $p_t$, of an English sentence. The semantic generator component consists of the valuation function to map the $p_t$ to its meaning. To accomplish this, we have introduced Semantic Knowledge Generation algorithm (Algorithm \ref{algorithm1}). First, the algorithm collects the list of verbs mentioned in the sentence and for each verb it accumulates all the syntactic (frame syntax) and corresponding semantic information (thematic roles and predicates)  from VerbNet using the verb-class of the verb. The algorithm finds the grounded thematic-role variables by doing a partial tree matching (described in Algorithm \ref{algorithm2}) between each gathered frame syntax and $p_t$. From the verb node of $p_t$, the partial tree matching algorithm performs a bottom-up search and, at each level through a depth-first traversal, it tries to match the skeletal parse tree of the frame syntax. If the algorithm finds an exact match or a partial match (by skipping words, e.g., prepositions), it returns the thematic roles to the parent Algorithm \ref{algorithm1}. Finally, Algorithm \ref{algorithm1} grounds the pre-defined predicate with the values of thematic roles and generates the ASP code.

\begin{algorithm}[h]
    \small
    \caption{Semantic Knowledge Generation}\label{algorithm1}
    \hspace*{\algorithmicindent} \textbf{Input:} \textit{$p_t$}: \text{constituency parse tree of a sentence}\\ 
    \hspace*{\algorithmicindent} \textbf{Output:} \textit{semantics}: \text{sentence semantics} 
    \begin{algorithmic}[1]
        \Procedure{GetSentenceSemantics}{$p_t$}
        \State $ \textit{verbs} ~ \gets  ~ \textit{getVerbs($p_t$)}$ \Comment{returns list of verbs present in the sentence}
        \State $ \textit{semantics} ~  \gets ~ \textit{\{\}}$ \Comment{initialization}
        \State \textbf{for each} {$v ~ \in ~ \textit{verbs}$} \textbf{do}
            \State \hspace{4mm} $ \textit{classes} ~ \gets ~ \textit{getVNClasses(v)}$ \Comment{get the VerbNet classes of the verb}
            \State \hspace{4mm} \textbf{for each} {$c ~ \in ~ \textit{classes}$} \textbf{do}
                \State \hspace{10mm}$ \textit{frames} ~ \gets ~ \textit{getVNFrames(c)}$ \Comment{get the VerbNet frames of the class}
                \State \hspace{10mm} \textbf{for each} {$f ~ \in ~ \textit{frames}$} \textbf{do}
                    \State \hspace{15mm} $ \textit{thematicRoles} ~ \gets ~ \newline \textit{getThematicRoles($p_t$, f.syntax, v)}$  \Comment{see Algorithm \ref{algorithm2}}
                    \State \hspace{15mm} $ \textit{semantics} ~ \gets ~ \textit{semantics} \; \; \;  \cup \; \; \;  \newline \textit{getSemantics(thematicRoles, f.semantics)}$ \\ \Comment{map the thematic roles into the frame semantics}
                \State \hspace{10mm} \textbf{end for}
            \State \hspace{4mm} \textbf{end for}
        \State \textbf{end for}
        \State \Return{semantics}
        \EndProcedure
    \end{algorithmic}
\end{algorithm}

\begin{algorithm}[h]
    \small
    \caption{Partial Tree Matching}\label{algorithm2}
    \hspace*{\algorithmicindent} \textbf{Input:} \textit{$p_t$}: \text{constituency parse tree of a sentence}; \textit{s}: \text{frame syntax}; \textit{v}: \text{verb}\\
    \hspace*{\algorithmicindent} \textbf{Output:} \textit{tr}: \text{thematic role set} or \textit{empty-set}: \text{\{\}} 
    \begin{algorithmic}[1]
        \Procedure{GetThematicRoles}{$p_t$, s, v}
        \State $ \textit{root} ~ \gets ~ \textit{getSubTree(node(v), $p_t$)}$  \Comment{returns the sub-tree from the parent of the verb node}
        \While{root}
           \State $tr ~ \gets ~ getMatching(root, s)$  \Comment{if s matches the tree return thematic-roles, else \{\}} 
           \If {$tr ~\neq ~\textit{\{\}}$} \Return tr  
            \EndIf
            \State $ \textit{root} ~ \gets ~  \textit{getSubTree(root, $p_t$)}$  \Comment{returns \textit{false} if root equals $p_t$}
        \EndWhile
        \State \Return{\{\}}
        \EndProcedure
    \end{algorithmic}
\end{algorithm}

The ASP code generated by the above mentioned approach represents the meaning of a sentence comprised of an action verb.
Since VerbNet does not cover the semantics of the \textit{`be'} verbs (i.e., \textit{am, is, are, have,} etc.), for sentences containing \textit{`be'} verbs, the \textit{semantic generator} uses pre-defined handcrafted mapping of the parsed information (i.e., syntactic parse tree, dependency graph, etc.) to its semantics. Also, this semantics is represented as ASP code.
The generated ASP code can now be used in various applications, such as natural language QA, summarization, information extraction, CA, etc.


\section{The SQuARE System}
Question answering system for reading comprehension is a challenging task for the NLU research community. In recent times with the advancement of ML applied to NLU, researchers have created more advance QA systems that show outstanding performance in QA for reading-comprehension tasks. However, for these high performing neural-networks based agents, the question rises whether they really \textit{``understand''} the text or not. These systems are outstanding in learning data patterns and then predicting the answers that require shallow or no reasoning capabilities. Moreover, for some QA task, if a system claims that it performs equal or better than a human in terms of accuracy, then the system must also show human level intelligence in explaining its answers. Taking all this into account, we have created our SQuARE QA system that uses ML based parser to generate the syntax tree and uses Algorithm \ref{algorithm1} to translate a sentence into its knowledge in ASP. By using the ASP-coded knowledge along with pre-defined generic commonsense knowledge, SQuARE outperforms other ML based systems by achieving 100\% accuracy in 18 tasks (99.9\% accuracy in all 20 tasks) of the bAbI QA dataset (note that the 0.01\% inaccuracy is due to the dataset's flaw, not of our system). SQuARE is also capable of generating English justification of its answers. 

\begin{figure}[h]
    \centering
    \includegraphics[scale = 0.7]{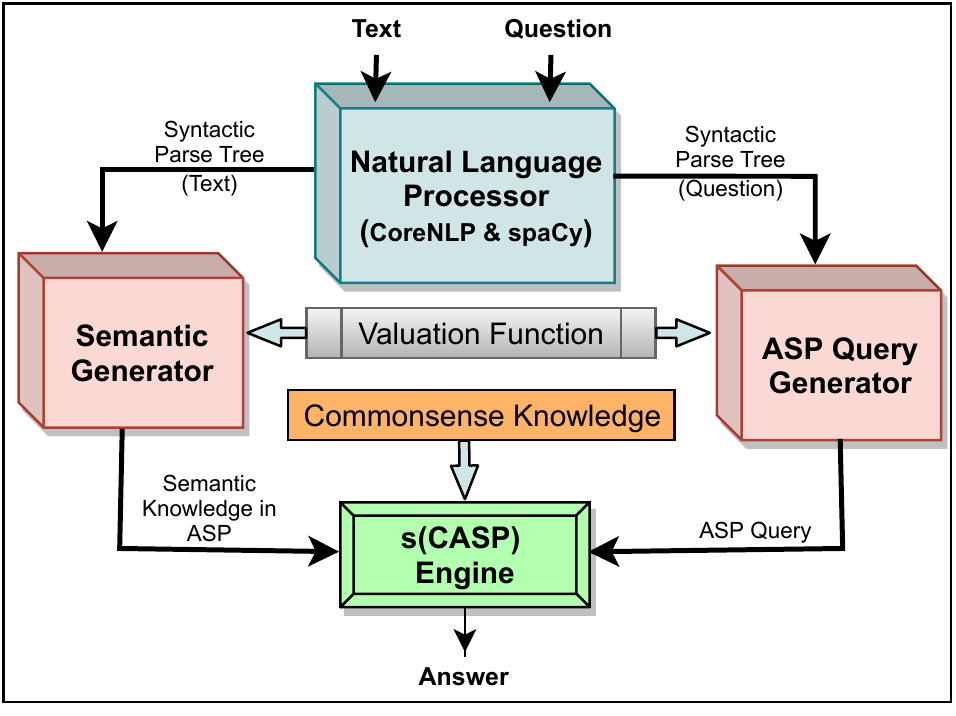}
    \caption{ SQuARE Framework }
    \label{fig:genral_framework}
    \end{figure}
\noindent\textbf{\underline{Architecture:}}
SQuARE is composed of two main sub systems: the \textit{semantic generator} and the \textit{ASP query generator}. Both subsystems inside the SQuARE architecture (illustrated in Figure \ref{fig:genral_framework}) share the common valuation function. To parse the passage and the question asked, the CoreNLP and spaCy parsers are used to generate the syntactic parse tree as well as the necessary parsing information such as NER, POS tags, lemmas, etc. Our semantic generator employs the process described earlier for ASP code generation from English sentences. The ASP query generator uses the same semantic generation algorithm with minor changes to adapt to the fact that the sentence is a question. It identifies the query variable along with the question type and other necessary details (e.g., NER, POS, dependency graph, etc.) from the parsed question, and then ASP query is formulated. For a given task, a  general query rule is defined that is a collection of sub-queries needed to answer the question. Such a rule can be considered as the process-template that a human will follow to achieve the same task, thus such process templates are a part of commonsense knowledge. To get the answer, the s(CASP) executes the query against the knowledge generated from the semantic-generator and the pre-defined, reusable commonsense knowledge.

SQuARE is also capable of reasoning over time using the order of the events in the passage. A passage or a story is a collection of sentences and the meaning of the whole passage comprises of meanings of the individual sentences. Knowledge is represented using defaults, exceptions and preferences which permits non-monotonic reasoning, needed because conclusion drawn early in a story may have to be revised later. SQuARE assumes the events in the story occur sequentially unless it encounters an exception. 

\medskip 
\noindent\textbf{\underline{Commonsense Concepts:}}
Just like humans, SQuARE uses predefined commonsense knowledge to understand the context. The dataset independent commonsense knowledge is written using generic predicate names so that it can be reused without any changes. This knowledge is presented either as facts or as default rules with exceptions. The facts are concrete global truth of the world, such as \texttt{property(color, white).} Whereas, default rules capture the normal relationship among two entities in the world or the general laws of the world, such as the law-of-inertia. Using negation as failure (NAF), these rules capture exceptions to defaults as well. NAF helps us to reason even if information is missing. Following default rule (used in task 20) illustrates that a thirsty person normally drinks unless there is an exception (e.g., a medical test needs to be performed which requires person does not drink any fluid in last 1 hour).

\smallskip 
\noindent    
\cprotect \fbox{
        \centering
        \begin{minipage}{0.94\linewidth}
        {\small
          \begin{verbatim}
action(X,drink) :- person(X),  emotional_state(X,
             thirsty),  not ab_action(X,drink).\end{verbatim}  
        }      
        \end{minipage}
    }
    
\smallskip\noindent 
Similarly, to have rudimentary human reasoning capabilities, SQuARE employs basic commonsense computational rules for counting objects, filtering, searching, etc. An example is given in the next section.  

\noindent\textbf{\underline{Example:}}
To demonstrate the power of the SQuARE system, we next discuss a full-fledged example showing the data-flow and the intermediate results.

\noindent \textit{\textbf{Story:} } A customized segment of a story from the bAbI QA dataset about \textit{counting objects} (Task-7) is taken.

\smallskip 
\noindent
\cprotect \fbox{
        \centering
        \begin{minipage}{0.94\linewidth}
        {\small
          \begin{verbatim}
1 John moved to the bedroom.
2 John got the football there.
3 John grabbed the apple there.
4 John picked up the milk there.
5 John gave the apple to Mary.
6 John left the football.\end{verbatim}  
        }      
        \end{minipage}
    }
    
\smallskip    
\noindent \textit{\textbf{Parsed Output: }} CoreNLP and spaCy parsers parse each sentence of the story and passes the parsed information to the semantic generator. Details are omitted due to lack of space, however, parsing can be easily done at \texttt{https://corenlp.run/}.

\noindent \textit{\textbf{Semantics: }} 
From the parsed information, the \textit{semantic generator} generates the semantic knowledge in ASP. We only give a snippet of knowledge (due to space constraint) generated from the third sentence of the story (VerbNet details of the verb - \textit{grab} is given in Figure \ref{fig:verbnet-example}).

\smallskip 
\noindent
\cprotect \fbox{
        \centering
        \begin{minipage}{0.94\linewidth}
        {\small
          \begin{verbatim}
1 contact(t3,during(grab),agent(john),
                               theme(the_apple)).
2 cause(t3,agent(john),event(grab)).
3 transfer(t3,during(grab),theme(the_apple)).\end{verbatim}  
        }      
        \end{minipage}
    }

\smallskip

\noindent \textit{\textbf{Question and ASP Query: } }
For the question - \textit{``How many objects is John carrying?''}, the\textit{ ASP query generator} generates a generic query-rule and the specific ASP query (it uses the process template for counting).

\smallskip
\noindent
\cprotect \fbox{
        \centering
        \begin{minipage}{0.94\linewidth}
        {\small
          \begin{verbatim}
count_object(T,Per,Count) :- 
    findall(O, property(possession,T,Per,O), Os), 
    set(Os,Objects), list_length(Objects,Count).
?- count_object(t6,john,Count). \end{verbatim}  
        }      
        \end{minipage}
    }

\smallskip 
\noindent \textit{\textbf{Answer: }} The s(CASP) finds the correct answer - \textit{1}

\noindent \textit{\textbf{Justification: }}
The s(CASP) generated justification for this answer is shown below: 

\smallskip

\noindent
\hbox {\hspace{-0.8em}\includegraphics [scale = 1.12]{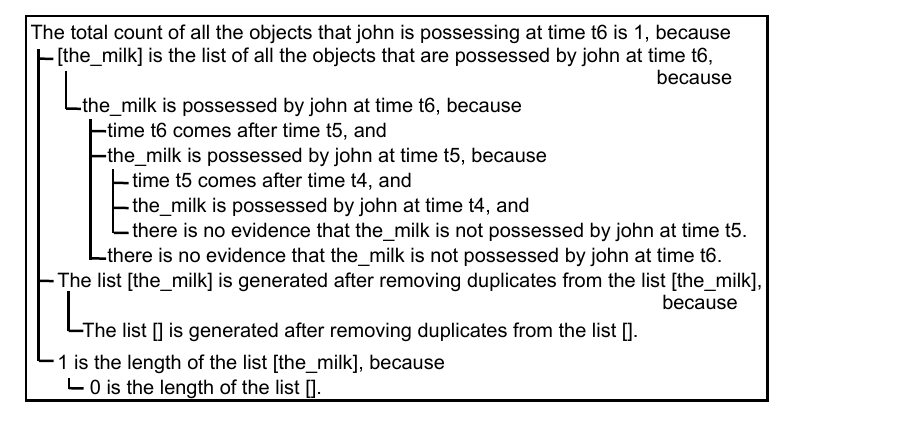}}

\section{StaCACK Framework} 
Conversational AI has been an active area of research, starting from a rule-based system, such as ELIZA \cite{eliza} and PARRY \cite{parry}, to the recent open domain, data-driven CAs like Amazon's Alexa, Google Assistant, or Apple's Siri. Early rule-based bots were based on just syntax analysis, while the main challenge of modern ML based chat-bots is the lack of ``understanding'' of the conversation. 
A realistic socialbot should be able to understand and reason like a human. In human to human conversations, we do not always tell every detail, we expect the listener to fill gaps through their commonsense knowledge. Also, our thinking process is flexible and non-monotonic in nature, which means \textit{``what we believe today may become false in the future with new knowledge''}. We can model this human thinking process with (i) default rules, (ii) exceptions to defaults, and (iii) preferences over multiple defaults \cite{asp}.

\begin{figure}[h]
    \centering
    \includegraphics[scale = 0.73]{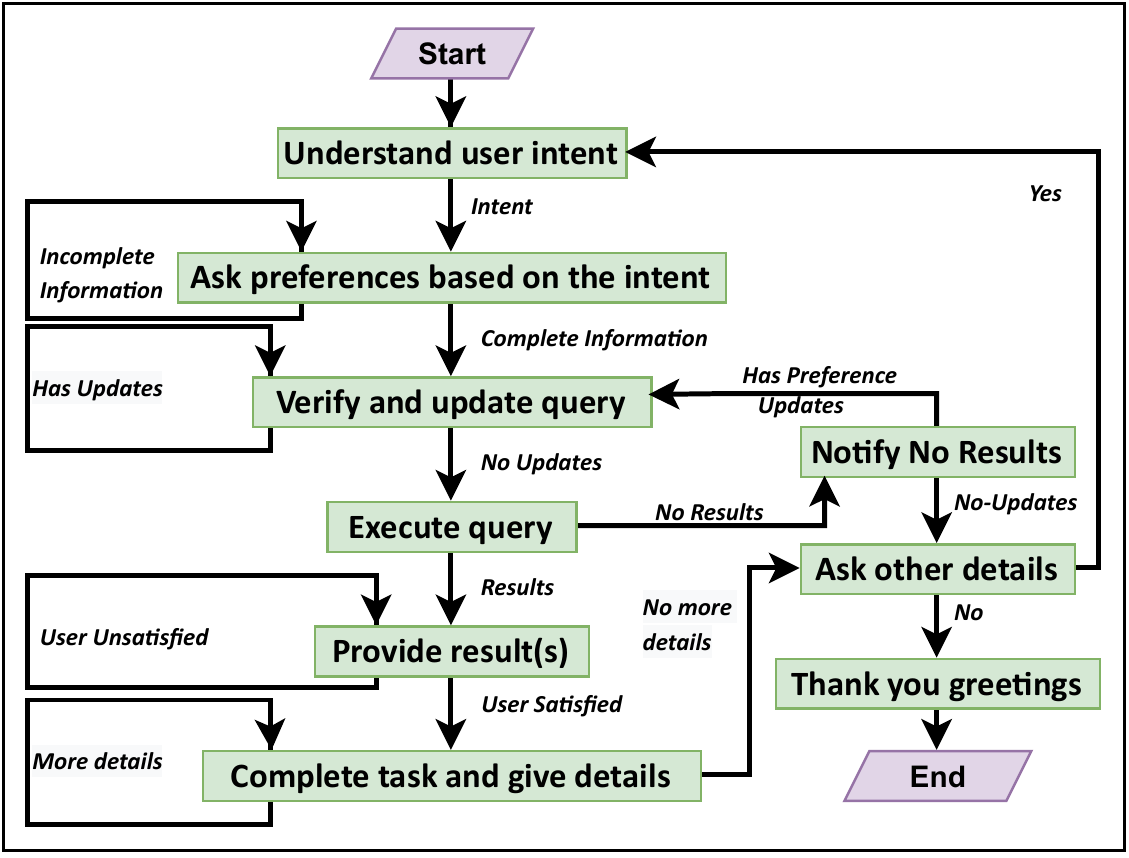}
    \caption{FSM for StaCACK framework}
    \label{fig:FSM_StaCACK}
\end{figure}

Following the discussion above, we have created StaCACK, a general closed-domain chatbot framework. StaCACK is a stateful framework that maintains states by remembering every past dialog between the user and itself. The main difference between StaCACK and the other stateful or stateless chatbot models is the use of commonsense knowledge for understanding user utterances and generating responses. Moreover, it is capable of doing non-monotonic reasoning by using defaults with exception and preferences in ASP. StaCACK achieves 100\% accuracy on the Facebook bAbI dialog dataset suit \cite{babi_dialog} (including OOV: out-of-vocabulary datasets) of five tasks  created for a restaurant reservation dialog system. In addition, StaCACK can answer questions that ML chatbots cannot without proper training (details are given in following sections). We focus on agents that are designed for a specific tasks (e.g., restaurant reservation).

\noindent\textbf{\underline{Finite State Machine:}}
Task-specific CAs follow a certain scheme in their inquiry that can be modeled as a finite state machine (FSM).
The FSM is illustrated in Figure \ref{fig:FSM_StaCACK}. However, the tasks in each state transition are not simple as in every level it requires different types of (commonsense) reasoning. We have tested our StaCACK framework on bAbI dialog dataset that deals with only one user intent - \textit{restaurant table reservation}. The whole end-to-end conversation, starting from understanding the user-intent to completing the reservation, is divided into four different tasks of the dataset - (a) issuing API calls (to actually make the reservation), (b) updating API calls, (c) Displaying options, and, (d) providing extra information. Therefore, if a system can complete these tasks, then it should be able to hold the whole conversation, that is given in the fifth dataset. However, using StaCACK, the implementation of the agent for these tasks become straightforward. We need to add commonsense knowledge about the domain (discussed later). Parsing of the sentences is done with spaCy and CoreNLP parsers and the parse tree is translated into knowledge using our Algorithm \ref{algorithm1}.

\noindent\textbf{\underline{Commonsense knowledge:}}
Similar to the SQuARE system, StaCACK augments knowledge generated from user utterances with commonsense knowledge. For a task based, general purpose CA, the commonsense knowledge or the background knowledge helps in understanding user-intent, missing information, and user preferences. For example, following rules illustrate how an agent can find all the missing parameter's value in the knowledge base, which later on can be used to ask counter-questions to the user to facilitate reasoning.  

\noindent
\cprotect \fbox{
        \centering
        \begin{minipage}{0.94\linewidth}
        {\small
          \begin{verbatim}
missing_parameter(X) :- query_parameter(X), 
        not query_parameter_value(X,_).
all_missing_parameter(ParaList) :-  
        findall(P,missing_parameter(P),ParaList).\end{verbatim}  
        }      
        \end{minipage}
    }

Due to the basis of StaCACK in commonsense knowledge, its reasoning can be more sophisticated and it can hold  conversation in a more natural manner (described in the next section) than ML based systems. Following is an example. Rule-1 depicts that normally a person who has consumed a type of cuisine (e.g., Mexican) the day before, would not like to eat it again today. However, there can be exceptions to this rule.

\noindent
\cprotect \fbox{
        \centering
        \begin{minipage}{0.94\linewidth}
        {\small
          \begin{verbatim}
1 cuisine_exception(P, C) :- property(person, P),
    property(cuisine, C), time(yesterday,P,C), 
    not ab_cusine_exception(P,C).
2 cuisine_suggestion(P,C) :- property(cuisine,C), 
    not cuisine_exception(P, C), 
    not ab_cuisine_suggestion(P,C).\end{verbatim}  
        }      
        \end{minipage}
    }

\smallskip 
\noindent\textbf{\underline{Example:}}
As we have already mentioned that the StaCACK framework shows excellent performance on the bAbI dialog dataset based on restaurant reservation conversation. Also, it is able to hold the conversation in a more natural way by using commonsense knowledge, which may not be possible with a rule-based system based on a monotonic logic. Following example shows how StaCACK can understand the cuisine preference of a user, just by performing reasoning over commonsense information about a cuisine (that \textit{curry} is predominant in \textit{Indian} and \textit{Thai} cuisine).

\smallskip 
\noindent
\cprotect \fbox{
        \centering
        \begin{minipage}{0.94\linewidth}
        {\small
          \begin{verbatim}
User:    Good morning	
StaCACK: Hello what can i help you with today
User:    Can you make a restaurant reservation in
         London in a cheap price range
StaCACK: I'm on it
User:    <SILENCE>	
StaCACK: any preference on a type of cuisine
User:    anything, except Lebanese food	
StaCACK: do you want to have Chinese food
User:    I want to have curry
StaCACK: Do you like Indian or Thai
User:    Thai\end{verbatim}  
        }      
        \end{minipage}
    }

\smallskip 
\noindent
 Please note that due to the space constraint, we have shown a conversation snippet here and how the end-to-end conversation looks like can be found elsewhere \cite{babi_dialog}.

\begin{table*}[b]
\centering
\tiny
\setlength{\belowcaptionskip}{0.5pt}
\begin{tabular}{|l||*{9}{c|}}
\hline
\backslashbox[13em]{\hphantom{thequ}Tasks}{Attributes\hphantom{th}} & 
    \makebox[2em]{\textbf{\cellcolor{blue!25}\begin{tabular}[c]{@{}c@{}}No. of \\ Stories\end{tabular}}} &
    \makebox[4em]{\textbf{\cellcolor{blue!25}\begin{tabular}[c]{@{}c@{}}Tot. No. of \\ Questions\end{tabular}}} &
    \makebox[6em]{\textbf{\cellcolor{blue!25}\begin{tabular}[c]{@{}c@{}}Avg. Questions \\ per Story\end{tabular}}} &
    \makebox[4em]{\textbf{\cellcolor{blue!25}\begin{tabular}[c]{@{}c@{}}Tot. No. of \\ Sentences\end{tabular}}} &
    \makebox[7em]{\textbf{\cellcolor{blue!25}\begin{tabular}[c]{@{}c@{}}Avg. Story Size\\ (No. of Sentences)\end{tabular}}} & 
    \makebox[7em]{\textbf{\cellcolor{blue!25}\begin{tabular}[c]{@{}c@{}}Max Story Size\\ (No. of Sentences)\end{tabular}}} &
    \makebox[3em]{\textbf{\cellcolor{blue!25}\begin{tabular}[c]{@{}c@{}}Accuracy\\ (\%)\end{tabular}}} & 
    \makebox[6em]{\textbf{\cellcolor{blue!25}\begin{tabular}[c]{@{}c@{}}Avg. Time \\ per Question\\ (Seconds)\end{tabular}}}\\\hline\hline
\textbf{\cellcolor{green!20}Single Supporting Facts} & 2200 & 11000 & 5 & 22000 & 10 & 10 & 100 & 15.4  \\\hline
\textbf{\cellcolor{green!20}Two Supporting Facts} & 2200 & 11000 &  5 & 48390 & 22 & 88 & 100 & 42.8\\\hline
\textbf{\cellcolor{green!20}Three Supporting Facts} & 2200 & 11000 & 5 & 163496 & 74 & 320 & 100 & 147.1\\\hline
\textbf{\cellcolor{green!20}Two Argument Relations} & 11000 & 11000  & 1 & 22000 & 2 & 2 & 100 & 0.6\\\hline
\textbf{\cellcolor{green!20}Three Argument Relations} & 2200 & 11000 & 5 & 59822 & 27 & 126 & 99.8 & 0.8\\\hline
\textbf{\cellcolor{green!20}Yes/No Questions} & 2200 & 11000 & 5 & 22662 & 10 & 26 & 100 & 2.1\\\hline
\textbf{\cellcolor{green!20}Counting} & 2200 & 11000 & 5 & 28514 & 13 & 52 & 100 & 3\\\hline
\textbf{\cellcolor{green!20}Lists/Sets} & 2200 & 11000 & 5 & 29394 & 13 & 58 & 100 & 5\\\hline
\textbf{\cellcolor{green!20}Simple Negation}  & 2200 & 11000 & 5 & 22000 & 10 & 10 & 100 & 1.2\\\hline
\textbf{\cellcolor{green!20}Indefinite Knowledge} & 2200 & 11000 & 5 & 22000  & 10 & 10 & 98.2 & 1.2\\\hline
\textbf{\cellcolor{green!20}Basic Coreference} & 2200 & 11000 & 5 & 22000 & 10 & 10  & 100 & 0.5\\\hline
\textbf{\cellcolor{green!20}Conjunction} & 2200 & 11000  & 5 & 22000 & 10 & 10 & 100 & 0.6\\\hline
\textbf{\cellcolor{green!20}Compound Coreference} & 2200 & 11000 & 5 & 22000 & 10 & 10 & 100 & 0.6\\\hline
\textbf{\cellcolor{green!20}Time Reasoning} & 2200 & 11000  & 5 & 25756 & 12 & 14 & 100 & 0.6\\\hline
\textbf{\cellcolor{green!20}Basic Deduction} & 2750 & 11000 & 4 & 22000 & 8 & 8 & 100 & 0.4\\\hline
\textbf{\cellcolor{green!20}Basic Induction} & 11000 & 11000 & 1 & 99000 & 9 & 9 & 100 & 0.8\\\hline
\textbf{\cellcolor{green!20}Positional Reasoning} & 1375 & 11000 & 8 & 2750 & 2 & 2 & 100 & 0.3\\\hline
\textbf{\cellcolor{green!20}Size Reasoning} & 2177 & 11000 & 5 & 13655 & 6 & 20 & 100 & 0.4\\\hline
\textbf{\cellcolor{green!20}Path Finding} & 11000 & 11000 & 1 & 55000 & 5 & 5 & 100 & 0.8\\\hline
\textbf{\cellcolor{green!20}Agent’s Motivations} & 1026 & 11000 & 11 & 11000 & 11 & 12 & 100 & 0.4\\\hline
\textbf{\cellcolor{orange!20}\textbf{Totals}} & \textbf{68,928} & \textbf{220,000} &  & \textbf{735,439} &  &  &  & \\\hline
\end{tabular}
\vspace{0.02in}
\caption{Dataset Statistics and Performance Results}
\label{table:square_performance_result}
\end{table*}

\section{Experiments and Results}

\noindent\textbf{\underline{Datasets:}}
The SQuARE and the StaCACK system have been tested on the bAbI QA \cite{babi} and the bAbI dialog dataset respectively \cite{babi_dialog}. With the aim of improving NLU research, Facebook researchers have created the bAbI datasets suit that comprise with different NLU application oriented simple task based datasets. The datasets are designed in such a way that it becomes easy for human to reason and reach an answer with proper justification whereas difficult for machines due to the lack of understanding about the language. 

These datasets are mainly created to train and test deep-learning based NLU applications. The bAbI QA dataset not only expects the ML models to train and test on the provided question answer pair for 20 reasoning based tasks but also presumes the model will learn the pattern of supporting facts given with each answer. Taking this into account, we choose this dataset to test SQuARE system to show its ``true understanding'' by generating natural language justification for each answer. Similarly, we have evaluated the StaCACK approach on the bAbI dialog dataset to exhibit how we can build a closed domain task based chatbot with commonsense knowledge. Another reason to choose these datasets is its simplicity that helps us to concentrate more on knowledge representation and modeling than pre-processing or parsing of English sentences.

\smallskip 
\noindent\textbf{\underline{Experiments:}}
In the SQuARE system, the accuracy has been calculated by matching the generated answer with the actual answer given in the bAbI QA dataset. Whereas, StaCACK's accuracy is calculated on the basis of \textit{per-response} as well as \textit{per-dialog}. Table \ref{table:square_performance_result} summarizes the testing statistics and performance metrics of the SQuARE system (benchmarks are tested on a \textit{intel i9-9900 CPU} with \textit{16G RAM}).
Table \ref{table:square_compare} and table \ref{table:Stacack_compare} compares our results in terms of accuracy with the existing state-of-the-art results for SQuARE and StaCACK system respectively.

\smallskip 
\noindent\textbf{\underline{Error Analysis:}}
SQuARE is not able to reach 100\% accuracy on two tasks  - the \textit{three argument relations} (Task 5) and the \textit{indefinite knowledge} (Task 10). We have studied the error stories and the particular questions where the system goes wrong. Thanks to the interpretable nature of the SQuARE system, we are able to identify the scenarios. 

\noindent
\textit{\textbf{Task 5}}: This error occurs because of the multiple correct answers present in the story 
with no indication of which one is the preferred one.
SQuARE system is capable of finding all the answers and all are correct, though bAbI (erroneously) assumes there is only one  correct answer. Illustration of the scenario is given below, where, clearly both \textit{milk} and \textit{apple} are answers, but bAbI only accepts \textit{milk}.

\smallskip 
    \noindent
\cprotect \fbox{
        \centering
        \begin{minipage}{0.94\linewidth}
        {\small
          \begin{verbatim}
1 Fred passed the milk to Bill.
2 Fred went back to the bedroom.
3 Fred took the apple there.
4 Fred gave the apple to Bill.
5 What did Fred give to Bill?   milk\end{verbatim}  
        }      
        \end{minipage}
    }

\medskip 
\noindent\textit{\textbf{Task 10}}: The erroneous stories have two identical places with an \textit{or} as a person's location. So, for the question about that person's location, SQuARE answers correctly, whereas the actual (erroneous) answer is \textit{`maybe'.} SQuARE correctly infers that $p \vee p$ equals $p$. Following example illustrates the scenario.
    
    \smallskip\noindent
\cprotect \fbox{
        \centering
        \begin{minipage}{0.94\linewidth}
        {\small
          \begin{verbatim}
1 Fred is either in the cinema or the cinema.
2 Is Fred in the cinema?    maybe\end{verbatim}  
        }      
        \end{minipage}
    }

\smallskip
\noindent
These errors show another advantage of our  explainable logic-based approach: SQuARE is able to identify errors in the dataset. On the contrary, the ML systems mimic the dataset by learning the errors as well and show their shallow understanding.

\begin{table}[b]
\centering
\tiny
\begin{tabular}{|c||*{3}{c|}}
\hline
\backslashbox[12em]{\hphantom{thequ}Tasks}{Model\hphantom{thequ}} & 
    \makebox[7em]{\textbf{\cellcolor{green!20}\begin{tabular}[c]{@{}c@{}}MemNN \\ (AM+NG+NL)\end{tabular}}} &
    \makebox[3em]{\textbf{\cellcolor{green!20}\begin{tabular}[c]{@{}c@{}}Mitra \\ et al.\end{tabular}}} &
    \makebox[3em]{\textbf{\cellcolor{green!20}\begin{tabular}[c]{@{}c@{}}SQuARE\end{tabular}}}\\\hline\hline
\textbf{\cellcolor{blue!25}Single Supporting Fact}  & 100 & 100 & 100 \\\hline
\textbf{\cellcolor{blue!25}Two Supporting Facts}  & 98 & 100 & 100 \\\hline
\textbf{\cellcolor{blue!25}Three Supporting Facts}  & 95 & 100 & 100 \\\hline
\textbf{\cellcolor{blue!25}Two Arg. Relation}  & 100 & 100 & 100 \\\hline
\textbf{\cellcolor{blue!25}Three Arg. Relation}  & 99 & 100 & 99.8 \\\hline
\textbf{\cellcolor{blue!25}Yes/No Questions}  & 100 & 100 & 100 \\\hline
\textbf{\cellcolor{blue!25} Counting}  & 97 & 100 & 100 \\\hline
\textbf{\cellcolor{blue!25}Lists/Sets}  & 97 & 100 & 100 \\\hline
\textbf{\cellcolor{blue!25}Simple Negation}  & 100 & 100 & 100 \\\hline
\textbf{\cellcolor{blue!25}Indefinite Knowledge} & 98 & 100 & 98.2 \\\hline
\textbf{\cellcolor{blue!25}Basic Coreference} & 100 & 100 & 100 \\\hline
\textbf{\cellcolor{blue!25}Conjunction} & 100 & 100 & 100 \\\hline
\textbf{\cellcolor{blue!25}Compound Coreference} & 100 & 100 & 100 \\\hline
\textbf{\cellcolor{blue!25}Time Reasoning} & 100 & 100 & 100 \\\hline
\textbf{\cellcolor{blue!25}Basic Deduction} & 100 & 100 & 100 \\\hline
\textbf{\cellcolor{blue!25}Basic Induction} & 99 & 93.6 & 100 \\\hline
\textbf{\cellcolor{blue!25}Positional Reasoning} & 60 & 100 & 100 \\\hline
\textbf{\cellcolor{blue!25}Size Reasoning} & 95 & 100 & 100 \\\hline
\textbf{\cellcolor{blue!25}Path Finding} & 35 & 100 & 100 \\\hline
\textbf{\cellcolor{blue!25}Agent’s Motivations} & 100 & 100 & 100\\\hline
\textbf{\cellcolor{orange!25}MEAN ACCURACY} & \textbf{94} & \textbf{100} & \textbf{100}\\\hline

\end{tabular}

\caption{SQuARE accuracy (\%)  comparison}
\label{table:square_compare}
\end{table}

\noindent\textbf{\underline{Comparison of Results and Related Works: }}
Using VerbNet to study the semantic relations of verbs and their semantic roles are not new. Text2DRS system \cite{text_to_drs} uses VerbNet to understand the discourse of a passage and represent it in the Neo-Davidsonian form. Schmitz et al. have also used VerbNet to design open information extraction system \cite{vn_ie}. Researchers have also used VerbNet to extend the commonsense knowledge about verbs. For instance, McFate \cite{cyc_vn} describes how VerbNet can be used to expand the CYC ontology by adding verb semantic frames. However all these approaches are limited to a particular domain or an application, and not scalable as well, whereas our semantic driven English sentence to ASP code generation approach is fully automatic process and independent of any application.

Table \ref{table:square_compare} compares our result in terms of accuracy with other models for all the 20 bAbI tasks for the SQuARE system. Note that due to the space constraint, we have shown comparison with two best performing systems (memory-neural-network (MemNN) with adaptive memory (AM), N-grams (NG), and non-linearity (NL); and the system by Mitra and Baral) and the accuracies on other models (i.e., N-gram classifier, LSTM, SVM, DMN, etc.) can be found elsewhere \cite{babi,dmn}. The SQuARE system beats all these ML based systems in terms of accuracy and explainability.

Mitra's system is the one closest to ours and motivated us to work on
the bAbI QA dataset. Their work achieves high accuracy, however, the
work is still a ML based inductive logic programming system that
requires manual annotation of data, such as mode declaration, finding
group size, etc. Conversely, the SQuARE system relies fully on
automatic reasoning with only manual encoding of \textit{reusable}
commonsense knowledge. An action language based QA methodology using
VerbNet has been developed by Lierler et al \cite{ALM}. The project
aims to extend frame semantics with ALM, \textit{an action language}
\cite{ALM}, to provide interpretable semantic annotations. Unlike
SQuARE, it is not an end-to-end automated QA system.

\begin{table}
\centering
\scriptsize
\setlength{\tabcolsep}{0.72em}
\begin{tabular}{|l|*{3}{r|}}
\hline
    &  \textbf{Mem2Seq}
      & \textbf{BossNet}
    & \textbf{StaCACK}
  \\[.82em]
  \hline\hline
\textbf{Task 1}  & 100 (100) & 100 (100)  & 100 (100) \\\hline
\textbf{Task 2}  & 100 (100) & 100 (100)  & 100 (100) \\\hline
\textbf{Task 3}  & 94.7 (62.1) & 95.2 (63.8)  & 100 (100) \\\hline
\textbf{Task 4}  & 100 (100) & 100 (100)  & 100 (100) \\\hline
\textbf{Task 5}  & 97.9 (69.6) & 97.3 (65.6)  & 100 (100) \\\hline
\textbf{Task 1 (OOV)} & 94.0 (62.2) & 100 (100)   & 100 (100)\\\hline
\textbf{Task 2 (OOV)} & 86.5 (12.4)  & 100 (100) & 100 (100) \\\hline
\textbf{Task 3 (OOV)} & 90.3 (38.7) & 95.7 (66.6)  & 100 (100) \\\hline
\textbf{Task 4 (OOV)} & 100 (100) & 100 (100) & 100 (100) \\\hline
\textbf{Task 5 (OOV)} & 84.5 (2.3) & 91.7 (18.5)  & 100 (100) \\\hline

\end{tabular}

\caption{Accuracy per response (per dialog) in \%.}
\label{table:Stacack_compare}
\vspace{-1em}
\end{table}

Table~\ref{table:Stacack_compare} shows the accuracy of our proposal,
StaCACK, against the best models on the bAbI dataset in terms of
per-response and in parenthesis in terms of per-dialog: Mem2Seq
\cite{mem2seq}, and BoSsNET \cite{bossnet}. Other results can be found
elsewhere~\cite{babi_dialog}.
Unsurprisingly, similar to the rule-based system, StaCACK surpasses
all the ML based models by showing 100\% accuracy. Nevertheless, due
to the commonsense reasoning, StaCACK can hold better natural
conversation (shown in the \textit{Example} section of
\textit{StaCACK}) that is not possible with a standard rule-based
system based on monotonic logic.

\section{Discussion}

Our goal is to create NLU applications that mimics the way human understand natural language. Humans understand a passage's meaning and use commonsense knowledge to logically reason to find a response to a question. We believe that this is the most effective process to create an NLU application. \textit{Learning} and \textit{reasoning} both are integral parts of human intelligence. Today, ML research dominates AI. Most state-of-the-art CA or QA systems have been developed using ML techniques (e.g., LSTM, GRU, Attention Network, Transformers, etc.). The systems that are built with these  techniques learn the patterns of the training text remarkably well and shows promising results on test data. With the recent advancements in the language model research, the pre-trained models such as BERT \cite{bert} and GPT-3 \cite{gpt3} have outstanding capability of generating natural languages. These rapid evolutions in the NLU world showcase some extremely sophisticated text predictor. These are used to build a chatbot or a QA system that can generate correct responses by exploiting the correlation among words and without properly understanding the content. These ML techniques are extremely powerful in tasks where learning of hidden data patterns is needed, such as machine translation, sentiment analysis, syntactic parsing, etc. However, they fail to generate proper responses where reasoning is required and they mostly do not employ commonsense knowledge. Also, the black-box nature of theses models makes their response non-explainable. In other words, these models do not possess any internal meaning representation of a sentence or a word and have no semantically-grounded model of the world. So, it will be an injustice to say that they understand their inputs and outputs in any meaningful way. Our semantic knowledge generation approach and its two applications are a step toward mimicking a human assistant. We believe that, to obtain truly intelligent behavior, ML and commonsense reasoning should work in tandem.

Compared to ML-based QA systems and CAs, our approach has many advantages. It produces correct responses by truly understanding the text and reasoning about it, rather than by using  patterns learned from training examples. Also, ML-based systems are more likely to produce incorrect response, if not trained appropriately, resulting in vulnerability under adversarial attacks \cite{ad_attack}. We believe that our commonsense reasoning based systems are more resilient. Scalability is also an issue due to the dependence on training data. Explainability is a necessary feature that a truly intelligent system must possess. Both SQuARE and StaCACK are capable of generating natural language justifications for the responses they produce.

\section{Future Work and Conclusion}
We presented our novel semantics-driven English text to answer set program generator. Also, we showed how commonsense reasoning coded in ASP can be leveraged to develop advanced NLU applications, such as SQuARE and StaCACK. We make use of the s(CASP) engine, a query-driven implementation of ASP, to perform reasoning while generating an natural language explanation for any computed answer. As part of future work, we plan to extend the SQuARE system to handle more complex sentences and eventually handle complex stories. Our goal is also to develop an open-domain conversational AI chatbot based on automated commonsense reasoning that can ``converse'' with a human based on ``truly understanding'' that person’s dialog.

\balance
\bibliographystyle{aaai.bst}
\bibliography{bibliography}

\end{document}